%% file: paper.tex
\relax
\documentclass[letterpaper]{article} 
\usepackage{aaai22}  
\usepackage{times}  
\usepackage{helvet}  
\usepackage{courier}  
\usepackage[hyphens]{url}  
\usepackage{graphicx} 
\usepackage{color}
\usepackage{natbib}  
\usepackage{caption} 
\DeclareCaptionStyle{ruled}{labelfont=normalfont,labelsep=colon,strut=off} 
\usepackage{algorithm}
\usepackage{algorithmic}

\usepackage{newfloat}
\usepackage{listings}
\floatstyle{ruled}
\newfloat{listing}{tb}{lst}{}
\floatname{listing}{Listing}

\usepackage{amsmath,amssymb}

\newcommand{\E}{\mathbb{E}}
\DeclareMathOperator*{\argmax}{argmax}

\title{Design Amortization for Bayesian Optimal Experimental Design}
\author{
    Noble Kennamer, \textsuperscript{\rm 1}
    Steven Walton, \textsuperscript{\rm 2}
    Alexander Ihler, \textsuperscript{\rm 1}
}
\affiliations{
    \textsuperscript{\rm 1}Department of Computer Science, University of California Irvine\\
   
    \textsuperscript{\rm 2}Department of Computer Science, University of Oregon\\
    nkenname@uci.edu
}

\begin{document}

\maketitle

\begin{abstract}

Bayesian optimal experimental design is a sub-field of statistics focused on developing methods to make efficient use of experimental resources. Any potential design is evaluated in terms of a utility function, such as the (theoretically well-justified) expected information gain (EIG); unfortunately however, under most circumstances the EIG is intractable to evaluate. In this work we build off of successful variational approaches, which optimize a parameterized variational model with respect to bounds on the EIG. Past work focused on learning a new variational model from scratch for each new design considered. Here we present a novel neural architecture that allows experimenters to optimize a single variational model that can estimate the EIG for potentially infinitely many designs. To further improve computational efficiency, we also propose to train the variational model on a significantly cheaper-to-evaluate lower bound, and show empirically that the resulting model provides an excellent guide for more accurate, but expensive to evaluate  bounds on the EIG. We demonstrate the effectiveness of our technique on generalized linear models, a class of statistical models that is widely used in the analysis of controlled experiments. Experiments show that our method is able to greatly improve accuracy over existing approximation strategies, and achieve these results with far better sample efficiency.

\end{abstract}

\section{Introduction}

Conducting experiments is often a resource-intensive endeavour, motivating experimenters to design their experiments to be maximally informative given the resources available. Optimal experimental design (OED) aims to address this challenge by developing approaches to define a utility, $U(d)$, of a possible designs, $d$, and algorithms for evaluating and optimizing this utility over all feasible designs $\mathcal{D}$. OED has been used widely across science and engineering, including systems biology \citep{boed_systems_biology}, geostatistics \citep{bayesian_geostatistical_design}, manufacturing \citep{oed_manufacturing} and more \citep{goos2011optimal}.

In this work we focus on evaluating the expected information gain (EIG), a commonly used utility function in Bayesian optimal experiment design (BOED) \citep{chaloner_1995_bayesian_oed_review, ryan_2016_BOED_algorithms_review}. We specify our model, composed of a likelihood and prior $p(y | \theta, d) p(\theta)$ for design $d$, possible experimental outcomes $y$ and latent variables $\theta$. The EIG is then defined to be:
\begin{equation}
    \label{eq:eig_def}
    EIG(d) = \E_{p(y|d)}\left[ H[p(\theta)] - H[p(\theta | y, d)]\right]
\end{equation}
where $H[\cdot]$ is the entropy function. The experimenter then seeks the solution to $\argmax_{d \in \mathcal{D}} EIG(d)$, where $\mathcal{D}$ is the set of all feasible designs. The EIG has sound theoretical justifications, proven to be optimal in certain settings \citep{sebastiani2000maximum, bernardo2009bayesian}.

While powerful, this framework is limited by the difficulty in evaluating the EIG due to the intractability of the posterior distribution $p(\theta |y, d)$. \citet{foster_vboed} proposed four variational bounds for efficiently approximating the EIG. The method involves defining a variational distribution, $q_{\phi}(\cdot)$ that will approximate either the posterior distribution $p(\theta |y, d)$ or the marginal likelihood $p(y | d)$. The parameters $\phi$ of this variational distribution are optimized according to the proposed bounds. In principle their method allows for estimating the EIG for arbitrarily complex models using flexible variational models, however their implementations focused on simpler variational forms that required fitting a new variational model for every possible design. In this work we focus on design amortization by proposing a novel deep learning architecture based on conditional normalizing flows (NF) \citep{papamakarios_2021_normalizing_flows, kobyzev_2020_normalizing_review} and set invariant models \citep{zaheer_deep_sets} to define a flexible variational distribution $q_{\phi}(\cdot | d)$ that only needs to be trained once, but can then accurately estimate the EIG for potentially infinitely many designs. Our experiments will show how design amortization can dramatically improve computational efficiency and how our more flexible variational form can make much more accurate approximations to the EIG over competing methods. We provide our code here\footnote{ redacted for anonymity during review.}.


\section{Background}
\label{sec:background}

Scientists consistently face the challenge of having to conduct experiments under limited resources, and must design their experiments to use these resources as efficiently as possible. BOED provides a conceptually clear framework for doing so. We assume we are given a model with design variables $d$, experimental outcomes $y$ and latent parameters $\theta$ about which we wish to learn. We have prior information on the latent variables encoded in a prior distribution, $p(\theta)$, and a likelihood that predicts experimental outcomes from a design and latent variables, $p(y | \theta, d)$. Via Bayes rule, these two functions combine to give us the posterior distribution $p(\theta |y, d) \propto p(y | \theta, d) p(\theta)$ representing our state of knowledge about the latent variables after conducting an experiment with design $d$ and observing outcomes $y$. For example the design variables, $d$, could represent the environmental conditions and chemical concentrations of a medium used to culture a strain of bacteria, which produces an important chemical compound.
This design problem becomes more complex with increasing dimension of $d$, for example, if we have $S$ petri dishes to work on (often called the \emph{experimental units} or \emph{subjects}). 
%
The experimental outcomes, $y$, would represent the amount of the chemical compound yielded from growing the culture in each of the conditions of $d$, and the latent variables $\theta$ represent parameters that define how the design variables $d$ mediate the yield of the chemical compound $y$. After conducting the experiment and observing $y$, we can quantify our information gain (IG) as:
%
\begin{equation}
\label{eq:IG}
\begin{split}
    IG(y, d) = H[p(\theta)] - H[p(\theta | y, d)]
\end{split}
\end{equation}
However, this gain cannot be evaluated before conducting the experiment, as it requires knowing the outcomes $y$. However, taking the expectation of the 
information gain 
with respect to the outcomes, $p(y|d)$, gives the EIG:
\begin{equation}
\label{eq:EIG}
\begin{split}
    EIG(d) &=\E_{p(\theta, y | d)}\left[\log\Big(\frac{p(\theta |y, d)}{p(\theta)}\Big)\right] \\
    &= \E_{p(\theta, y | d)}\left[\log\Big(\frac{p(y|\theta, d)}{p(y|d)}\Big)\right]
\end{split}
\end{equation}
\paragraph{Nested Monte Carlo:}
Typically $p(\theta |y, d)$ and $p(y|d)$ are intractable, making the EIG challenging to compute. One common approach to approximating EIG is to use a nested Monte Carlo (NMC) estimator \citep{nmc_tutorial_adaptive_design_optimization, nmc_vincent2017darc, rainforth_nestingMC}:
\begin{equation}
    \label{eq:nmc}
    \begin{split}
    \hat{\mu}_{NMC} = \frac{1}{N} \sum_{n=1}^{N} \log \frac{p(y_{n} | \theta_{n, 0}, d)}{\frac{1}{M}\sum_{m=1}^{M} p(y_{n} | \theta_{n, m}, d)} \\
    \text{ where } \theta_{n, m} \sim p(\theta)
    \text{ and } y_{n} \sim p(y| \theta_{n, 0}, d)
    \end{split}
\end{equation}
\citet{rainforth_nestingMC} showed that NMC is a consistent estimator converging as $N, M \xrightarrow[]{} \infty$. They also showed that it is asymptotically optimal to set $M \propto \sqrt{N}$, resulting in the overall convergence rate of $\mathcal{O}(T^{-\frac{1}{3}})$, where $T$ is the total number of samples drawn (i.e. $T=NM$ for NMC).  However, this is much slower than the $\mathcal{O}(T^{-\frac{1}{2}})$ rate of standard Monte Carlo estimators \citep{robert1999monte}, in which the total number of samples is simply $T=N$.

The slow convergence of the NMC estimator can be limiting in practical applications of BOED.  The inefficiencies can be traced to requiring an independent estimate of the marginal likelihood, $p(y_{n} | d)$, for each $y_{n}$ (the denominator of Eq.~\eqref{eq:nmc}). Inspired by this, \citet{foster_vboed} proposed employing techniques from variational inference by defining a functional approximation to either $p(\theta |y, d)$ or $p(y | d)$, and allowing these estimators to amortize across the samples of $y_{n}$ for more efficient estimation of the EIG. In this work we focus on two of the four estimators they proposed: the posterior estimator and variational nested Monte Carlo. 

\paragraph{Posterior Estimator:} The posterior estimator is an application of the Barber-Agakov bound to BOED, which was originally proposed for estimating the mutual information in noisy communication channels \citep{barber_agakov_bound}. It requires defining a variational approximation $q_{\phi}(\theta |y, d)$ to the posterior distribution, giving a lower bound to the EIG:
\begin{equation}
    \label{eq:posterior_estimator}
    \begin{split}
        EIG(d) \ge \mathcal{L}_{post}(d) &\triangleq \E_{p(\theta, y | d)} \left[\log\Big(\frac{q_{\phi}(\theta |y, d)}{p(\theta)}\Big)\right] \\
        &\approx \frac{1}{N} \sum_{n=1}^{N} \log \frac{q_{\phi}(\theta_{n} |y_{n}, d)}{p(\theta_{n})} \\
     \text{ where } y_{n}, \theta_{n} \sim p(y, \theta | d).\!\!\!\!\!\!\!\!\!\!\!\!\!\!\!\!\!
    \end{split}
\end{equation}
By maximizing this bound with respect to the variational parameters $\phi$, we can learn a variational form that can efficiently estimate the EIG. A Monte Carlo estimate of this bound converges with rate $\mathcal{O}(T^{-\frac{1}{2}})$, and if the true posterior distribution is within the class of functions defined by the variational form $q_\phi$, the bound can be made tight (dependent on the optimization) \citep{foster_vboed}.


\paragraph{Variational Nested Monte Carlo:} The second bound we discuss is variational nested Monte Carlo (VNMC). It is closely related to NMC, but differs by applying a variational approximation $q_{\phi}(\theta |y, d)$ as an importance sampler to estimate the marginal likelihood term in NMC:
%
\begin{multline}    
    \label{eq:vnmc_estimator}
        EIG(d) \le \\ \mathcal{U}_{VNMC}(d, M) \triangleq 
        \E \left[ \log \frac{p(y | \theta_{0}, d) }{\frac{1}{M} \sum_{m=1}^{M} \frac{p(y, \theta_{m} | d)} {q_{\phi}( \theta_{m} | y, d)}} \right]
\end{multline}
where the expectation is taken with respect to $y, \theta_{0:M} \sim p(y, \theta_{0} | d) \prod_{m=1}^{M} q_{\phi}( \theta_{m} | y, d)$.

By minimizing this upper bound with respect to the variational parameters $\phi$, we can learn an importance distribution that allows for much more efficient computation of the EIG. Note that if $q_{\phi}(\theta |y, d)$ exactly equals the posterior distribution, the bound is tight and requires only a single nested sample ($M=1$). Even if the variational form does not equal the posterior, the bound remains consistent as $M \xrightarrow{} \infty$.  Finally, it is worth noting that by taking $q_\phi(\theta|y,d) = p(\theta)$, the estimator simply reduces to NMC.

It was further shown by \citet{foster_unified} that VNMC can be easily made into a lower bound by including $\theta_{0}$ (the sample from the prior) when estimating the marginal likelihood, a method we denote as contrastive VNMC (CVNMC):
\begin{multline}
    \label{eq:contrastive_vnmc_estimator}
        EIG(d) \ge \\ \mathcal{L}_{CoVNMC}(d, M) \triangleq 
        \E \left[ \log \frac{p(y | \theta_{0}, d) }{\frac{1}{M+1} \sum_{m=0}^{M} \frac{p(y, \theta_{m} | d)} {q_{\phi}( \theta_{m} | y, d)}} \right]
\end{multline}
where the expectation is taken with respect to $y, \theta_{0:M} \sim p(y, \theta_{0} | d) \prod_{m=1}^{M} q_{\phi}( \theta_{m} | y, d)$.
We can also employ this same technique to regular NMC to estimate both lower and upper bounds.

Note that the upper bound \eqref{eq:vnmc_estimator} and lower bound \eqref{eq:contrastive_vnmc_estimator} are particularly useful when evaluating the performance of our method in settings where ground truth is not available. In these cases we can examine the bound pairs produced by NMC and by VNMC to assess which set more tightly constrains the true value.

\paragraph{Practical considerations.} 
In this work, we apply the same flexible mathematical framework proposed in \citet{foster_vboed}.  However,
\citet{foster_vboed} adopted a ``classical'' variational distribution setting, in which the variational form $q_\phi$
is selected to take a standard, parametric form.  They found this approach effective, but tested only on very simple
design problems, with only one experimental unit at a time.
Their variational models only incorporate the design implicitly, requiring a separate optimization for every design
to be considered%
\footnote{Although subsequent work \citep{foster_unified} considered evolving both the design and 
distribution $q$ simultaneously,
even that work remains focused on a single (if evolving) design.}.
Unfortunately, as we show in the experiments
this approach is not effective on more complex design problems.
Instead,  
we propose a far more flexible, deep learning based distributional form that incorporates the design explicitly,
allowing us to amortize training across and apply our trained model to evaluation of all 
(potentially continuously many) designs in our feasible set.


\section{Method}
\label{sec:methods}
We are interested in learning a parameterized function, $q_{\phi}( \theta | y, d)$, for approximating the posterior distribution. We now describe our proposed deep learning architecture for amortizing over designs, allowing practitioners to train a single model that is capable of evaluating the EIG for potentially infinitely many designs. We also discuss how we can efficiently train this model using the (simpler and cheaper) equation \eqref{eq:posterior_estimator}, then use the resulting approximation in the more accurate bounds provided by VNMC, \eqref{eq:vnmc_estimator}--\eqref{eq:contrastive_vnmc_estimator}. This advances the work in \citet{foster_vboed} by providing a highly flexible variational form that can be used in a wide variety of contexts and an inexpensive procedure to train it.

\subsection{Neural Architecture}
\begin{figure}[t]
    \centering
    \includegraphics[width=0.45\textwidth]{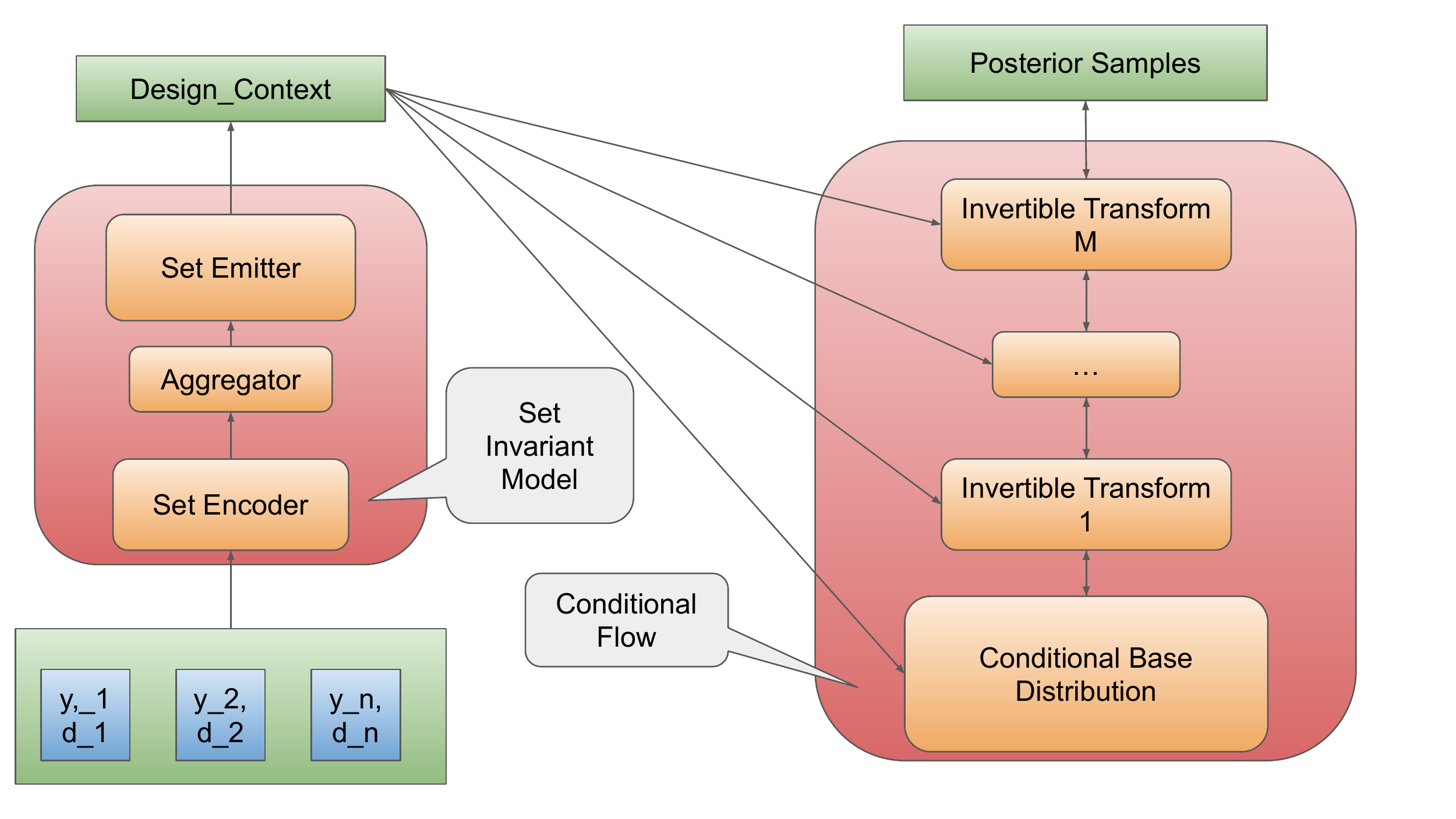}
    \caption{A high-level schematic of our architecture for amortizing over designs. The first component (left) takes in the design variables and simulated observations and produces a design context, $c_{y,d}$.  In many experiments the individual units being experimented on are exchangeable, thus we use a set invariant architecture.  The second (right) is a conditional normalizing flow, conditioned on the design context produced by the first component. Together, they define our variational posteriors $q_{\phi}(\theta |y, d)$, amortized over designs.}
    \label{fig:neural_arch}
\end{figure}

Figure \ref{fig:neural_arch} shows a high level representation of our architecture. Broadly, it consists of two major components. The first is a learnable function for taking in the design variables $d$ and simulated experimental outcomes from the model, $y$, and producing a design context vector, $c_{y,d}$, that will be used to define a conditional distribution. We focus on the common case where the experimental units 
lack any meaningful order 
and our learnable function must therefore be permutation invariant. We can incorporate this inductive bias into our model by making our function follow the general form of set functions proposed in \citet{zaheer_deep_sets}. In the sequel, we denote this component as our \emph{set invariant model}. The second major component is a learnable distribution conditioned on the design context produced by the set invariant model. In this work we use conditional normalizing flows, which consist of a base distribution and sequence of invertible transformations with tractable Jacobian determinants to maintain proper accounting of the probability density through the transformations. Both the base distribution and transformations are learnable and conditioned on the design context. 

\newcommand{\emit}{{\textsc{Emit}}}
\newcommand{\encode}{{\textsc{Enc}}}

\paragraph{Set Invariant Model.} It is often the case that the individual units being experimented on do not posses an inherent ordering -- for example, 
subjects in a randomized controlled clinical trial, or
the petri dishes in 
our previous example.  
Suppose we would like to find the optimally informative design for an experiment with $S$ experimental units, where $d_{i}$ and $y_{i}$ denote the design variables and simulated outcomes of unit $i$, respectively.  In this setting we want our design context to be invariant to permutations in its inputs, e.g., reordering the individuals in the trial should not change our results. Learning permutation invariant functions is an active area of research 
\citep[e.g.,][]{invariant_nn_bloem2020probabilistic}. In this work we follow the general form proposed by \citet{zaheer_deep_sets}, 
where our set invariant model is defined as,
\begin{equation}
    \label{eq:set_arch}
       c_{y,d} = \emit_{\phi_{\emit}}\left[ \sum_{i=0}^{S} \encode_{\phi_{\encode}}(y_{i}, d_{i})  \right].
\end{equation}
In particular, we define two learnable functions. The set encoder $\encode_{\phi_{\encode}}(y_{i}, d_{i})$ takes as input the design variables and simulated outcomes for each individual experimental unit. Its output is an intermediary representation for each experimental unit, which are aggregated together by summation; the permutation invariance of the sum ensures invariance of the overall function. The aggregated representation is then passed through the set emitter function $\emit_{\phi_{\emit}}(\cdot)$, which creates the final design context used in the conditional normalizing flow. In our experiments we find substantially improved performance using attention layers \citep{attention_is_all_you_need} in the set encoder, which allows for interactions between the experimental units before aggregation. In this case we should denote our set encoder as $\encode_{\phi_{encode}}(y_{i}, d_{i} | y_{-i}, d_{-i})$ where $y_{-i}$ $d_{-i}$ denote all simulated outcomes and design variables except for the $i\text{th}$ unit.

Structuring our design encoder function in this way gives two major advantages. First, permutation invariance does not need to be learned by the function since it is already present by construction; this can make learning more efficient and reduces the total number of weights via weight sharing. Second, the function is able to encode designs with a variable number of experimental units, $S$, as long as the $d_{i}$ and $y_{i}$ have the same size for all units.

Note that not all experimental design problems are permutation invariant in the experimental units. For example, in some settings there could be a temporal component in the design variables, in which case we could replace our set invariant function with an order-based model such as a recurrent neural network.

\paragraph{Conditional Normalizing Flow.} Normalizing flows define an expressive class of learnable probability distributions, which has been used in generative modeling and probabilistic inference \citep{papamakarios_2021_normalizing_flows, kobyzev_2020_normalizing_review}. The main idea of normalizing flow based models is to represent a random variable $\theta$ as a transformation $\theta = T(x)$ of a random variable $x$ sampled from a base distribution $p(x)$. The key property is that the transformation $T$ must be invertible and differentiable. This allows us to obtain $p(\theta)$ via a change of variables,
\begin{equation}
    \label{eq:change_of_variables}
       p(\theta) = p(x) \  |\det J_{T}(x)|^{-1}
\end{equation}
where $x = T^{-1}(\theta)$ and $\det J_{T}(x)$ is the determinant of the Jacobian at $x$. Both the transformations and base distribution may have learnable parameters.  This provides a highly flexible class of distributions that can be both sampled and efficiently evaluated.
%

In our setting we would like to learn not just a single distribution, but rather a \emph{conditional distribution} given design variables and experimental outcomes. This conditioning on $d$ is key to allowing us to amortize over all possible designs. 
To this end, we learn a sequence of $K$ conditional transformations $T_{\phi_{i}}(\cdot | c_{y,d})$ and a conditional base distribution $p_{\phi_{0}}(x| c_{y,d})$ which together define our variational approximation to the posterior distribution amortized over designs.
\begin{equation}
    \label{eq:post_nf}
        q_{\phi}(\theta |c_{y,d}) = 
       p_{\phi_{0}}(x_{0}|c_{y,d})\prod_{i=1}^{K}|\det J_{T_{\phi_{i}}}(x_{i} | c_{y,d})|^{-1}
\end{equation}
%
%
%
where $\theta = T_{\phi_{K}} \circ T_{\phi_{K-1}} \circ \ldots T_{\phi_{1}} (x_{0})$, with all transformations and base distribution conditioned on the design context \eqref{eq:set_arch}. The full architecture, including the set invariant model and conditional normalizing flow is trained end-to-end.

\subsection{Variational Posterior Training}
\label{sec:variational_posterior_training}

The posterior estimator and the (contrastive) VNMC bounds all require learning a variational approximation to the posterior distribution. In both \citet{foster_vboed} and \citet{foster_unified} each bound was trained separately, learning its own variational approximation. However in all cases the variational approximation produced by training one of the bounds \emph{can} be used for evaluating any other bound, since they all only require an approximate posterior distribution. Ideally, we would like to train using only the posterior estimator since it is much cheaper -- a total cost of only $\mathcal{O}(N)$ -- whereas both VNMC bounds have a total cost of $\mathcal{O}(NM)$. 
However, it is not obvious that training on the (also less accurate) bound should still
provide good EIG estimates when used in the VNMC bounds. 
Our experiments show that it is surprisingly effective across a broad range of models. 
Intuitively speaking, this is possible because all bounds share the same optimum -- the true posterior distribution. 
Moreover, because the posterior estimator takes its expectation with respect to the model $p(y, \theta |d)$, the variational approximation $q_{\phi}(\theta |y, d)$ will in general be \emph{wider} than the true posterior distribution, akin to variational inference using the ``forward'' Kullback-Liebler (KL) divergence,
and in contrast to the more commonly used ``reverse KL'' variational optimization methods that result in underdispersed and mode-seeking optima.
This property also makes the posterior estimator's $q_\phi$ an excellent choice for importance sampling (as in VNMC), in which 
a too-narrow proposal distribution can lead to high variance in the importance weights, causing a small number of samples to dominate the
estimator \citep[Chapter. 9]{owen_mcbook}.
%

\section{Related Work}
\label{sec:related_work}
Our approach builds on the framework for BOED developed in \citet{foster_vboed}, which proposed four variational bounds for estimating EIG.  The framework itself is quite flexible, capable of accommodating a wide variety of models (e.g., implicit vs explicit likelihoods), sequential experimentation (of arbitrary batch size) and arbitrary variational forms $q$.  However, their experiments were limited to 
only single experimental units,
and used simple variational forms that cannot amortize over designs (requiring a separate training procedure for each proposed design). 
In this work we propose a deep learning architecture which can easily be scaled to approximate arbitrarily complex distributions. In addition, our architecture can amortize over designs, allowing us to train a single variational model capable of estimating the EIG for potentially infinitely or continuously many designs. We also show that we can train our model using the cheaper posterior bound, then use its optimized approximate posterior within the VNMC bounds for a more accurate final approximation. We show that, using our proposed variational form, 
we can achieve highly accurate EIG estimates across a spectrum of complex design problems. 
While a few other EIG approximations have been proposed (see, e.g., \citet{foster_vboed, ryan_2016_BOED_algorithms_review}), in light of the
experimental results of \citet{foster_vboed} we mainly compare our experimental performance relative to NMC.
%


\section{Experiments}
\label{sec:experiments}
We perform three types of experiments: \emph{amortization}, \emph{model} and \emph{architecture} experiments.
Our \emph{amortization} experiment shows the dramatic increase in efficiency from amortization, and better EIG estimation provided by our more complex variational forms compared to those used in \citet{foster_vboed}. \emph{Model} experiments examine how the benchmark method, NMC, breaks down as model complexity grows while our methods remain reliable for accurately estimating the EIG.  \emph{Architecture} experiments measure the impact of key componets in our variational approximation and serve as a guide to using our method effectively. 

In all experiments we focus on estimating designs for different types of generalized linear models (GLMs) \citep{mccullagh_1989_glm}. GLMs are a very common model class used to analyze controlled experiments and are regularly used in applications of optimal experimental design \citep{goos2011optimal}. Our GLMs have the general pattern, 
\begin{equation}
    \label{eq:glm}
    \begin{split}
       &\theta \sim \mathcal{N}(\mu_{p}, \Sigma_{p}) \\
       &r = g^{-1}(D\theta )\\
       &y \sim \text{Exponential Family Distribution} (r).
    \end{split}
\end{equation}
Here, $\theta$ is a $N_{p}+1$ dimensional parameter vector, where $N_{p}$ is the number of predictors ($+1$ for the intercept term). $D$ is a $N_{E} \times (N_{p}+1)$ design matrix, where $N_{E}$ is the number of experimental units. The inverse link function, $g^{-1}$, defines the type of GLM.  Finally, $\mu_{p}$ and $\Sigma_{p}$ are the prior mean and covariance of the parameters.  
%
Our experiments cover six GLMs: normal (known observation noise), normal unknown (unknown observation noise), logistic, binomial, categorical and multinomial. For the normal model with known observation noise we take $\sigma=1$; for the normal model with unknown observation noise%
\footnote{In this case, the observation noise is included as the standard deviation in the normal distribution that samples $y$.}
we use the prior $\sigma \sim \text{InverseGamma}(a_{p}, b_{p})$ with $a_{p}=b_{p}=3.5$. 
For the binomial model, we assume 10 random trials; we use 3 classes for the categorical model, and 10 trials and 3 classes for the multinomial model. All experiments in the sequel are run for designs with 5 experimental units. Our implementations made significant use of Pyro \citep{bingham_pyro_cite} to implement the inference procedures and NFlows \citep{nflows_software} to construct our conditional normalizing flows. All training was done on a single Nvidia 2080TI and evaluation was done on an Intel I7-9800X with 64 GB of RAM.

\subsection{Amortization Experiments}
\label{sec:amorization_experiments}
\begin{figure}[t]
    \centering
    \includegraphics[width=.9\columnwidth]{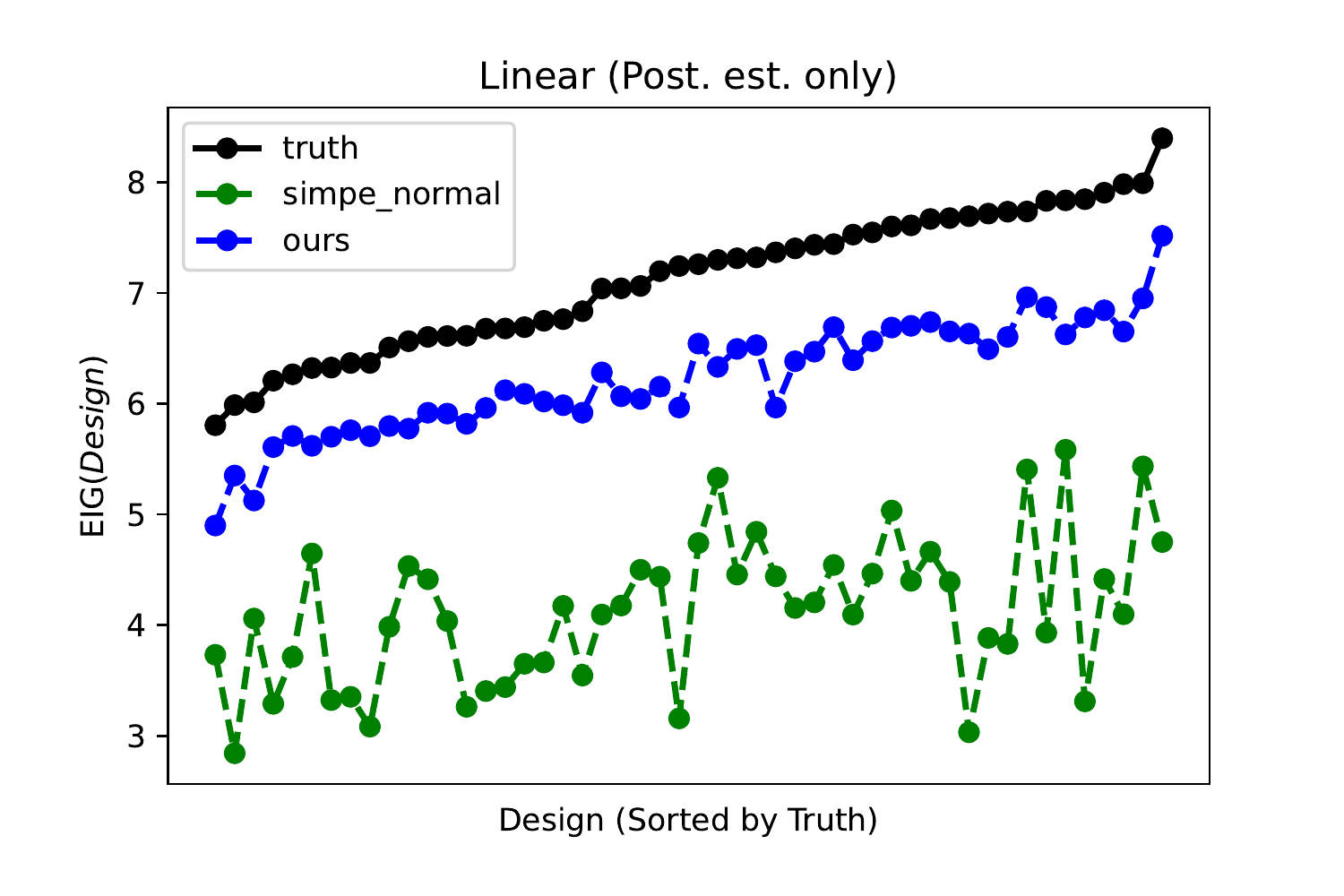}
    \caption{Comparing our method and the (non-amortized) variational form used by \citet{foster_vboed} on the normal (linear) model with 5 predictors and 5 on the diagonal of the prior covariance (similar to the AB model in \citet{foster_vboed}).  For clarity we only show the posterior estimator values. 
    Our method is $>3\times$ faster (293s vs.~920s) and significantly more accurate. See text for further analysis.}
    \label{fig:amortization_compare}
\end{figure}
While providing an excellent framework, the variational forms used in \citet{foster_vboed} are too simple to be effective on the GLM models we consider. Additionally, their work required training a new variational model for every design being approximated, while we propose a method that can amortize over designs.
Our closest model to those tested in \citet{foster_vboed} is our normal (linear) model, similar to their ``AB model''; we apply their variational
form for the AB model (given in their appendix) in order to perform a comparison.
We set $N_p=5$ and $\Sigma_p = 5 I$, i.e., diagonal with variance 5.

We generate 50 random designs with $N_E = 5$ experimental units and compare the quality of EIG approximations given by the posterior estimator as well as total wall clock time. The precise architecture and training procedure we use are described in the appendix. Figure \ref{fig:amortization_compare} shows the results of this experiment: our method produces a much tighter lower bound that is highly correlated with the true values; selecting the highest estimate would pick the design with highest true EIG in the set. In contrast, the variational form used in \citet{foster_vboed} yields a much looser and less correlated bound, which would select the 6th best design if used.  Moreover, our method is more than $3 \times$ faster (293 seconds compared to 920 seconds), showing the benefit of amortizing over designs. In fact, this speed-up understates how much more computationally efficient our method is, given that it leaves us with a model that can estimate the EIG for arbitrarily many designs without additional training.  Training took 291 of our method's 293 seconds; evaluating an additional 50 designs, then, would take virtually the same amount of time, compared to double the time required for the non-amortized approach.
The non-amortized training is prohibitively slow, 
while moreover for our other GLM models it is often not clear what variational form from \citet{foster_vboed} could be applied;
for these reasons, in the rest of the experiments we compare only to standard NMC.


\subsection{Model Experiments}
\label{sec:model_experiments}
In our model experiments we vary the GLMs in two ways: the number of predictors  (not including the intercept) and the diagonal components of the prior covariance (all off-diagonal terms are zero). The number of predictors $N_p$ is varied from 1 to 5 and the diagonal of the prior covariance in $\{1, 5, 25\}$. For the neural network architecture we use attention layers for the set encoder, a residual network for the set emitter \citep{he_resnet}, a full rank Gaussian distribution for the conditional base of the normalzing flow and four affine coupling layers each parameterized with a residual network. Additional details of the architecture and training parameters can be found in the appendix. In all experiments we train the variational approximation using the posterior estimator. During training, new designs are generated randomly from a multivariate normal distribution with identity covariance in dimension $N_p+1$. For final evaluation we generate 50 new random designs and estimate the posterior bound with $N=5000$ samples, while the VNMC bounds are estimated with $N=1000$ and $M = 31$ samples and nested samples, and the NMC bounds are estimated with $N=30000$ and $M = 173$ samples and nested samples. The number of samples for VNMC and NMC were selected based on the maximum number of samples that fit into memory (64 GB RAM) for the largest model (multinomial with 5 predictors).

\begin{figure}[t]
    \centering
    \includegraphics[width=0.99\columnwidth]{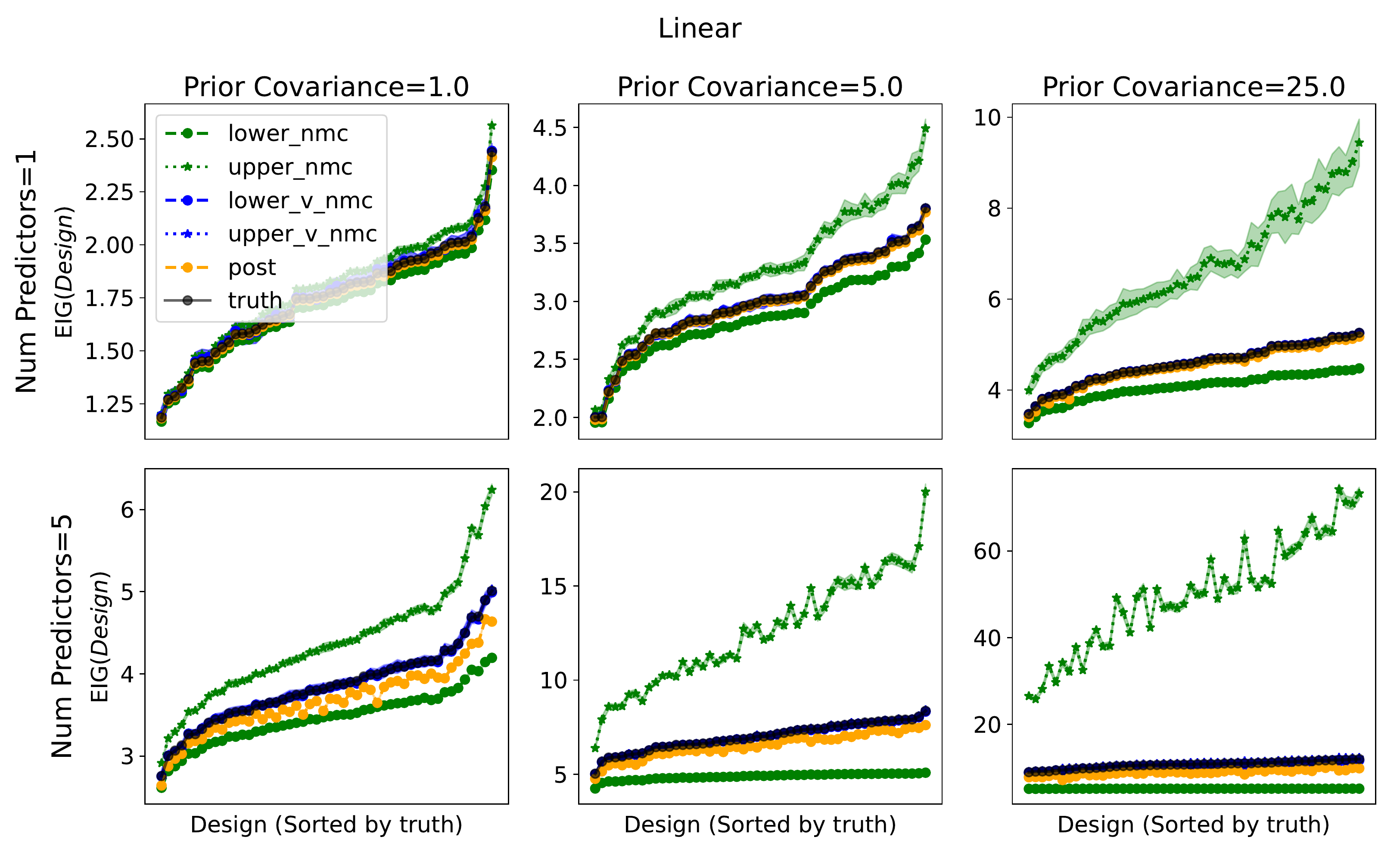}
    \caption{Results for estimating EIG in the linear model with known observation noise. The x-axis ranges over the index of 50 randomly selected designs, each with 5 experimental units. Due to the dimensionality, the designs lack a meaningful order; for visual clarity we plot them in the sorted order of the true EIG. The rows vary the number of predictors ($N_p\in \{1, 5\}$) while columns show changes in the diagonal of the prior covariance matrix, $\{1, 5, 25\}$, from informative to uninformative.  NMC and VNMC methods can estimate both upper and lower bounds, while the posterior estimator only provides a lower bound. 
    Our proposed method gives much tighter bounds on the truth than the competing NMC (with $167\times$ fewer samples). 
    The shading shows one standard deviation of our estimates over 20 runs.}
    \label{fig:linear_model}
\end{figure}

Figure \ref{fig:linear_model} shows EIG evaluations for 50 randomly generated designs for the 5 estimators: posterior, VNMC upper, VNMC lower (contrastive VNMC), NMC upper and NMC lower (contrastive NMC). Since this figure pertains to the linear model we can calculate the ground truth EIG exactly, shown in solid black.  For visual clarity we sort the designs in order of ground truth EIG value.  We see that all estimators perform reasonably well on the easiest form of the model (1 predictor with unit prior covariance). However, even in this case the VNMC bounds (upper and lower) more tightly constrain the ground truth -- in fact both are nearly exact. In addition the posterior bound (a lower bound) is consistently above the NMC lower bound and closer to the truth. These trends become magnified as the prior covariance and number of predictors increase. In all cases the VNMC bounds are nearly exact, while the performance of NMC degrades rapidly with problem difficulty.  Again, the posterior estimator remains above and closer to truth than the NMC lower bound. 

\begin{figure}[t]
    \centering
    \includegraphics[width=.99\columnwidth]{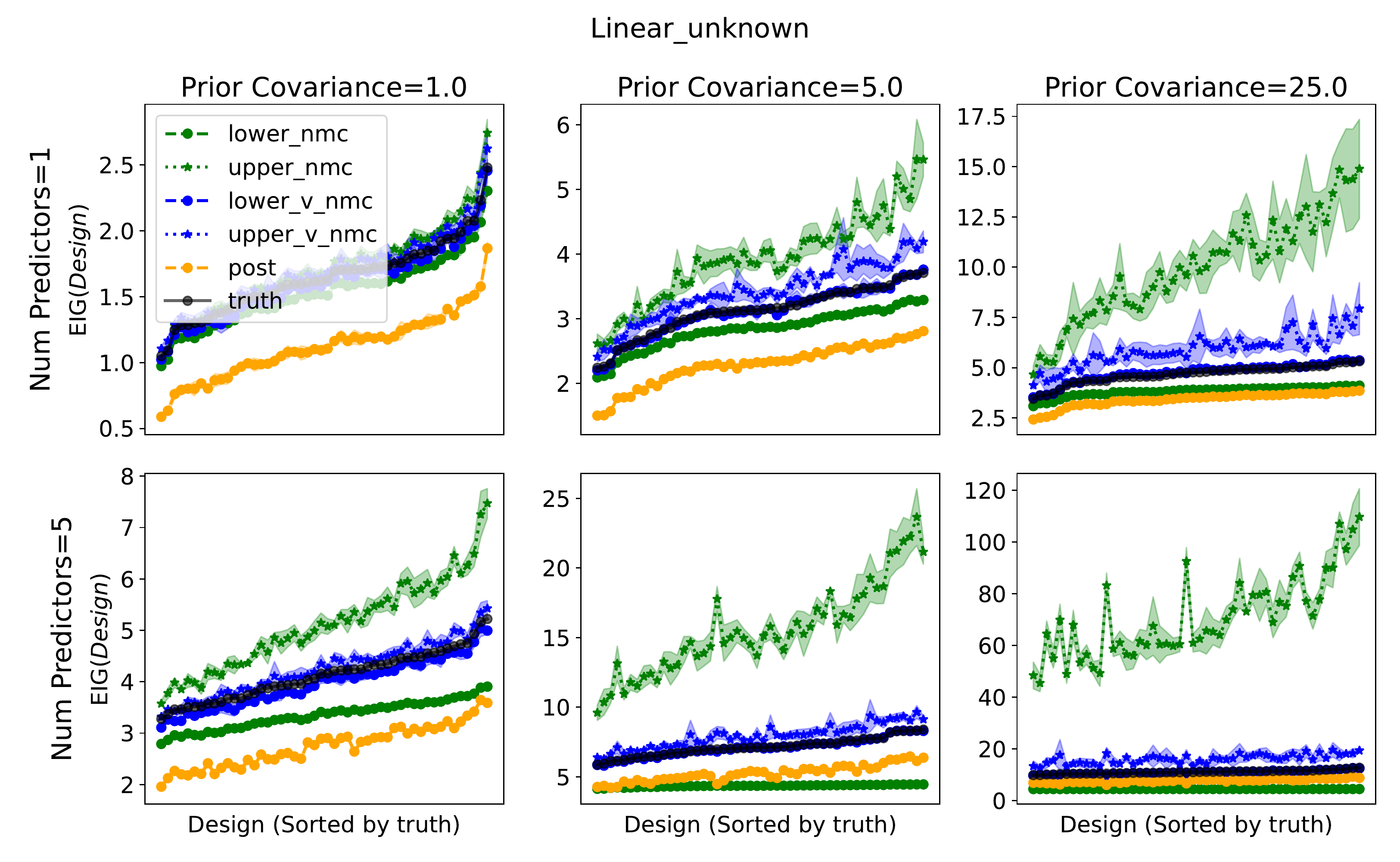}
    \caption{Same as Figure \ref{fig:linear_model}, but for the linear model with unknown observation noise.}
    \label{fig:linear_unknown_model}
\end{figure}

Figure \ref{fig:linear_unknown_model} shows exactly the same set of experiments, but for the linear unknown model. In this case we can calculate a high-quality Monte Carlo estimate of the ground truth thanks to conjugacy, and sort the designs to be evaluated in order of this ground truth. The results are largely consistent with those from the linear model with known observation noise: the VNMC bounds constrain the ground truth much more tightly than the NMC bounds. However in this case the posterior estimator is only above the NMC lower bound in the two hardest cases (5 predictors and 5 or 25 diagonal covariance). 

Figures \ref{fig:logistic_model}, \ref{fig:binomial_model}, \ref{fig:categorical_model}, \ref{fig:multinomial_model}, in the appendix, show the same set of experiments but for the logistic, binomial, categorical and multinomial models. In none of these cases can we calculate ground truth, so all plots order their designs by the benchmark NMC (upper). Even without ground truth we still clearly see the lower and upper bounds of VNMC are much closer together and below/above their NMC counterparts in practically all cases. One exception is the logistic model, where as long as the model only contains one predictor variable, the NMC bounds are as tight as the VNMC bounds; however by 5 predictors the VNMC bounds are tighter.  In fact the VNMC lower and upper bounds agree with each other in all cases, suggesting they are closely estimating the true EIG. We also see that the VNMC lower and upper bounds are touching across all settings of the binomial model and all but the hardest in the categorical and multinomial models (where they are still much tighter than NMC), again suggesting that our VNMC estimators are nearly exact. The results for the posterior estimator are more mixed -- sometimes it is above the NMC lower bound and sometimes below.
Nevertheless, the variational posterior learned using the posterior estimator can be used to compute VNMC bounds on the EIG that are much tighter than NMC.

Our experiments show that training a \emph{single} variational posterior, amortizing over designs, we can calculate the EIG much more accurately than the competing NMC benchmark, nearly calculating ground truth exactly.  Moreover, the experiments show that training on the posterior estimator can provide a variational distribution that remains effective for estimation using the more costly VNMC bounds (see Section \ref{sec:variational_posterior_training}). Not only does VNMC provide far more accurate estimates, it does so with many fewer samples -- more than two order of magnitudes ($167\times$) fewer samples than NMC. 


\subsection{Architecture Experiments}
\label{sec:arch_experiments}

We next investigate the importance of architectural decisions for the neural networks defining $q_{\phi}(\theta |y, d)$.  We compare using attention layers vs.\ residual layers in the set encoder, and transform type and number in the normalizing flow.  We compare 
using 4 vs.\ 8 transforms, and test
affine coupling transforms \citep{dinh_realNVP}, rational quadratic (RQ) splines \citep{durkan_spline_flows}, and no transform (just the conditional base distribution). 
We run all combinations for the linear unknown and binomial model with 5 predictors and 25 on the diagonal of the prior covariance.  Note that the true posterior for the linear unknown model is $t$-distributed, while the binomial is not analytically expressible, so we expect the use of normalizing flows to be advantageous over just the normal base distribution. The rest of the architecture components are the same as the Model Experiments and full details can be found in the appendix.

Figure \ref{fig:arch_loss_linear_unknown} shows the loss curves of the experiments for the linear unknown model. The inset plot on the top right shows the loss curves across all epochs, while the main plot shows a detail of the last 50 optimization steps. Each optimization step is run on a batch of 50 designs, so this plot indicates final performance on 2500 randomly generated designs. Specifically the loss is $-\sum_{i=1}^{N} \log q_{\phi}(\theta_{i} |y_{i}, d)$ where $y_{i}, \theta_{i} \sim p(y, \theta |d)$ and $N=50$ -- the cross entropy of the variational posterior.  Empirically, we see that using attention layers in the set encoder is the most important architectural decision (lower 5 curves vs upper); all networks using attention layers achieved superior performance to all networks using ResNets regardless of the other architectural settings. Beyond this, we see that using an affine coupling layer is also important, but see little difference between 4 and 8 transform layers. Surprisingly, the RQ transforms perform no better than having no transform. 
This is because the RQ transforms are restricted to the range of $(0, 1)$, with linear tails outside. Even after training, almost all parameter samples are outside this range by the time they reach the spline; the linear tails effectively skip the flow layers completely, explaining why its performance is comparable to models with no transform layers. 
In the appendix, Figure \ref{fig:arch_eig_linear_unknown} shows the performance of a subset of these models at evaluating the EIG for 50 random designs, highlighting that the loss curves' values are directly related to the accuracy and sample efficiency of the EIG estimate. Figures \ref{fig:arch_loss_binomial} and \ref{fig:arch_eig_binomial} show the same plots for the binomial model and further support these conclusions.

\begin{figure}[t]
    \centering
    \includegraphics[width=.99\columnwidth]{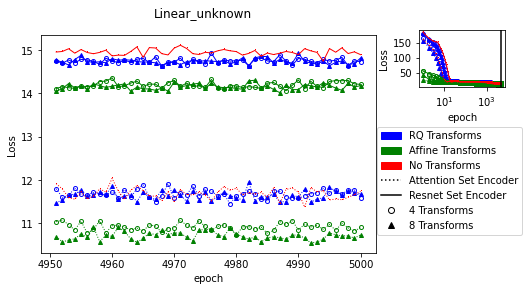}
    \caption{Results for our architecture experiments on the linear unknown model with 5 predictors and diagonal prior covariance 25.  We vary 
    the architecture of the set encoder (attention vs.~resnet), the normalizing flow transform type (affine coupling, affine spline, or no transform), and the number of transforms (4 or 8). The main plot shows the loss for the posterior estimator over the last 50 steps of training; each step is performed over a batch of 50 random designs (2500 designs total). The inset plot shows the loss curves over all 5000 training steps, indicating all architectures have converged. Further discussion is given in the text.}
    \label{fig:arch_loss_linear_unknown}
\end{figure}

\section{Conclusion}
\label{sec:conclusion}
In this paper we expand on the work of \citet{foster_vboed}, which proposed variational bounds for estimating the EIG for Bayesian optimal experimental design. In particular we propose a deep learning architecture incorporating set invariance and conditional normalizing flows that allows us to train a single model capable of estimating the EIG across the design space. Our experiments show that this architecture is highly effective at estimating EIG, and that design amortization provides significant computational speed ups.  For cases where ground truth can be calculated, our model's VNMC bounds are nearly exact, while in cases without ground truth our VNMC upper and lower bounds are often sufficiently tight to suggest they are exact.  These estimates are significantly more accurate than those of standard NMC while requiring far ($167\times$) fewer samples, as well as far more accurate and efficient than the simpler, non-amortized variational forms used in \cite{foster_vboed}.  We also demonstrate that we can train our model using the much cheaper posterior estimator bound, with cost $\mathcal{O}(N)$, then evaluate using this fitted model within the more accurate but costly VNMC bounds, $\mathcal{O}(NM)$.  Together, we provide a method for faster and more accurate approximation of the EIG across many possible designs.  In future work, we plan to extend our approach to design optimization tasks; on this point, we observe that our variational form, $q_{\phi}(\theta |y, d)$, is differentiable with respect to the designs, $d$, suggesting it can generalize and potentially improve
on the gradient-based design optimization objectives proposed by  \citet{foster_unified}.
%

\bibliography{paper} 


\begin{appendix}
\include{appendix}

\end{appendix}

\end{document}

%% file: appendix.tex
\section*{Appendix}

\subsection{Full Amortization Experiment Details}

The full details of the architecture we used for our method and training parameters are described in the Appendix under ``Full Architecture Details". For amortization comparison we adopted the variational form used for the AB model in \citep{foster_vboed} as it is the most similar to the model that we test on. The variational form is:
\begin{equation}
    \label{eq:simple_variational_form}
    \begin{split}
       q_{\phi}(\theta |y, d) = \mathcal{N}(Ay, \Sigma).
    \end{split}
\end{equation}
where the variational parameters are $\phi = \{A, \Sigma\}$ with $A$ being a $N_{E} \times (N_{p}+1)$ matrix and $\Sigma$ a $(N_{p}+1) \times (N_{p}+1)$ positive definite matrix.  As we can see this form incorporates the design $d$ only implicitly, and cannot amortize over designs. Recall that $N_{E}$ is the number of experimental units and $N_{p}$ is the number of predictors (adding 1 for the intercept). We train using the AdamW optimizer for 5000 steps with a learning rate of $.001$ with $\beta_{0}=.9$ and $\beta_{1}=.999$.

\subsection{Additional Model Experiment Results}
\begin{figure*}
    \centering
    \includegraphics[width=1.9\columnwidth]{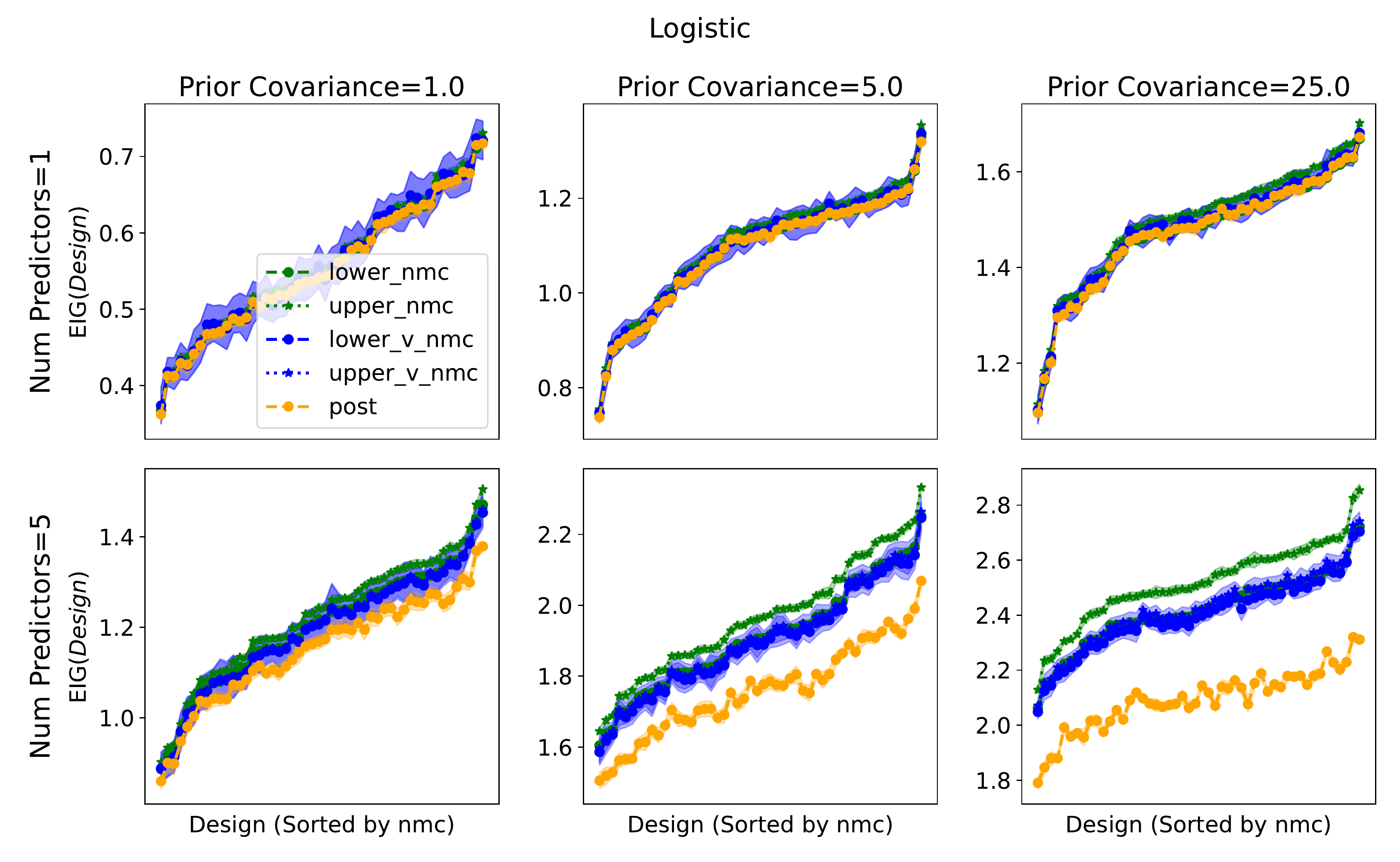}
    \caption{Same as figure \ref{fig:linear_model}, but for the logistic model. For this model we cannot compute ground truth, but we can still infer superior performance for our methods by observing how VNMC produces tighter bounds, both lower and upper compared to NMC. Designs are ordered on the x-axis via the baseline NMC. }
    \label{fig:logistic_model}
\end{figure*}

\begin{figure*}
    \centering
    \includegraphics[width=1.9\columnwidth]{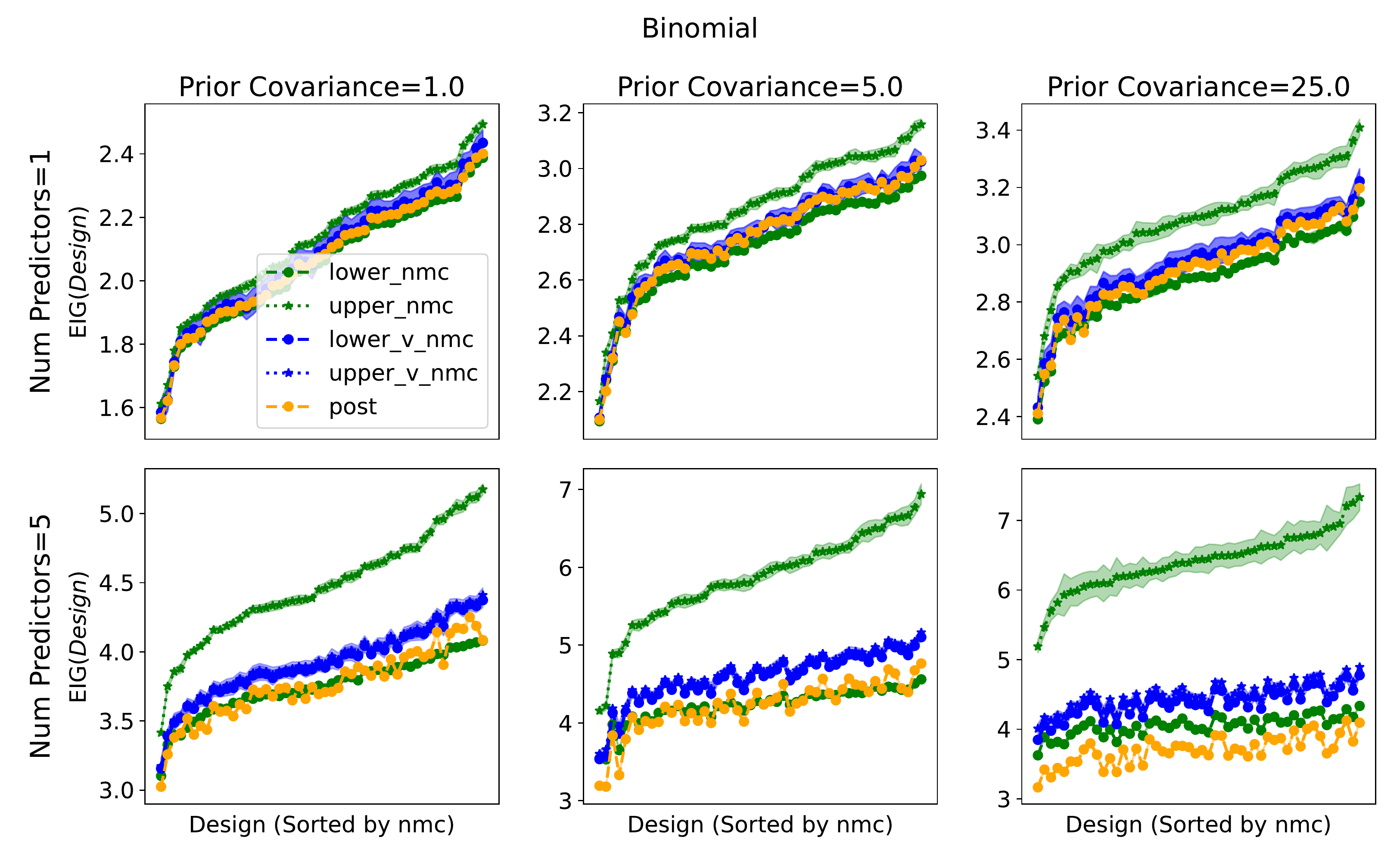}
    \caption{Same as figure \ref{fig:linear_model}, but for the binomial model. For this model we cannot compute ground truth, but we can still infer superior performance for our methods by looking at how VNMC produces much tighter bounds, both lower and upper.}
    \label{fig:binomial_model}
\end{figure*}

\begin{figure*}
    \centering
    \includegraphics[width=1.9\columnwidth]{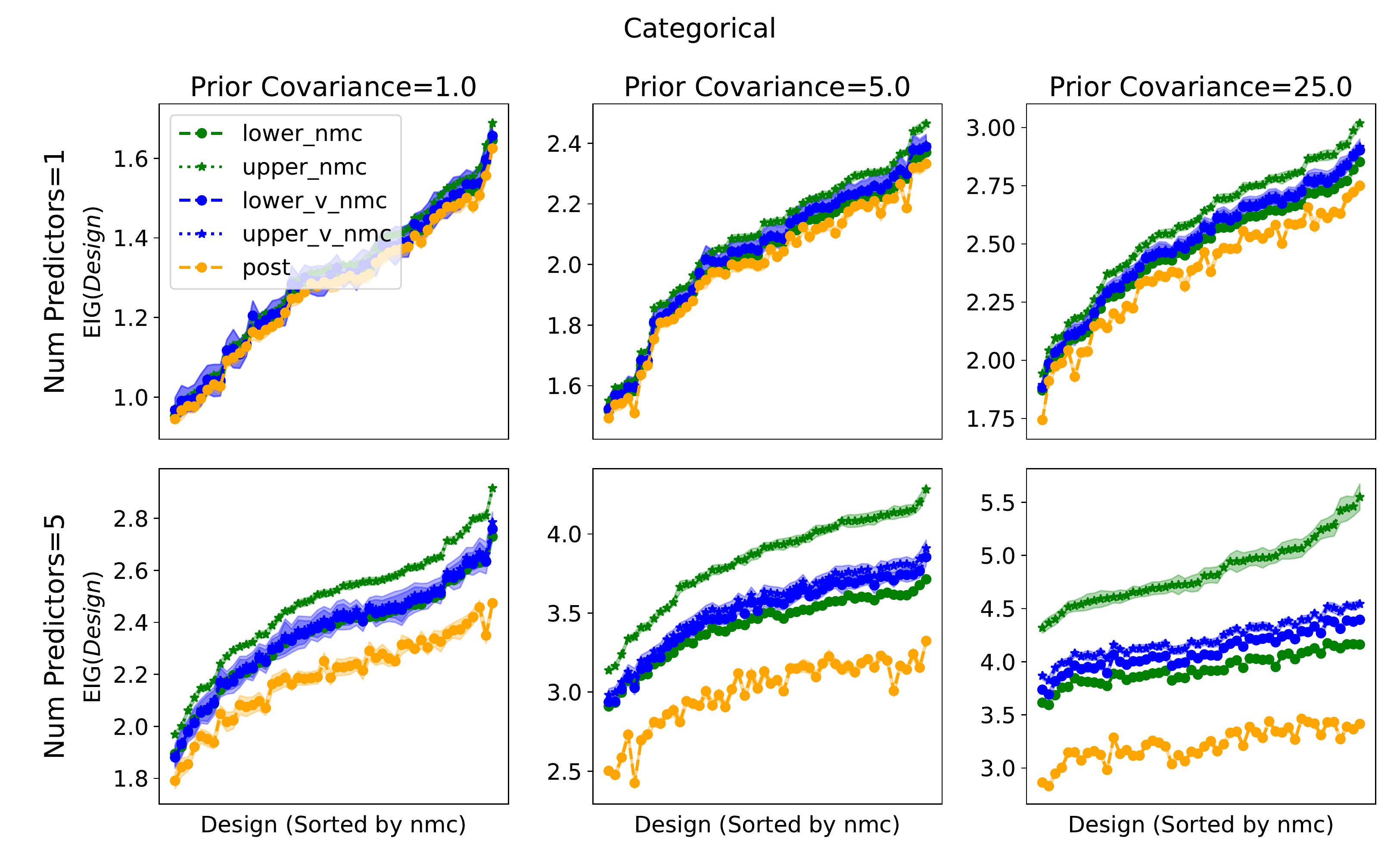}
    \caption{Same as figure \ref{fig:linear_model}, but for the categorical model. For this model we cannot compute ground truth, but we can still infer superior performance for our methods by looking at how VNMC produces much tighter bounds, both lower and upper.}
    \label{fig:categorical_model}
\end{figure*}

\begin{figure*}
    \centering
    \includegraphics[width=1.9\columnwidth]{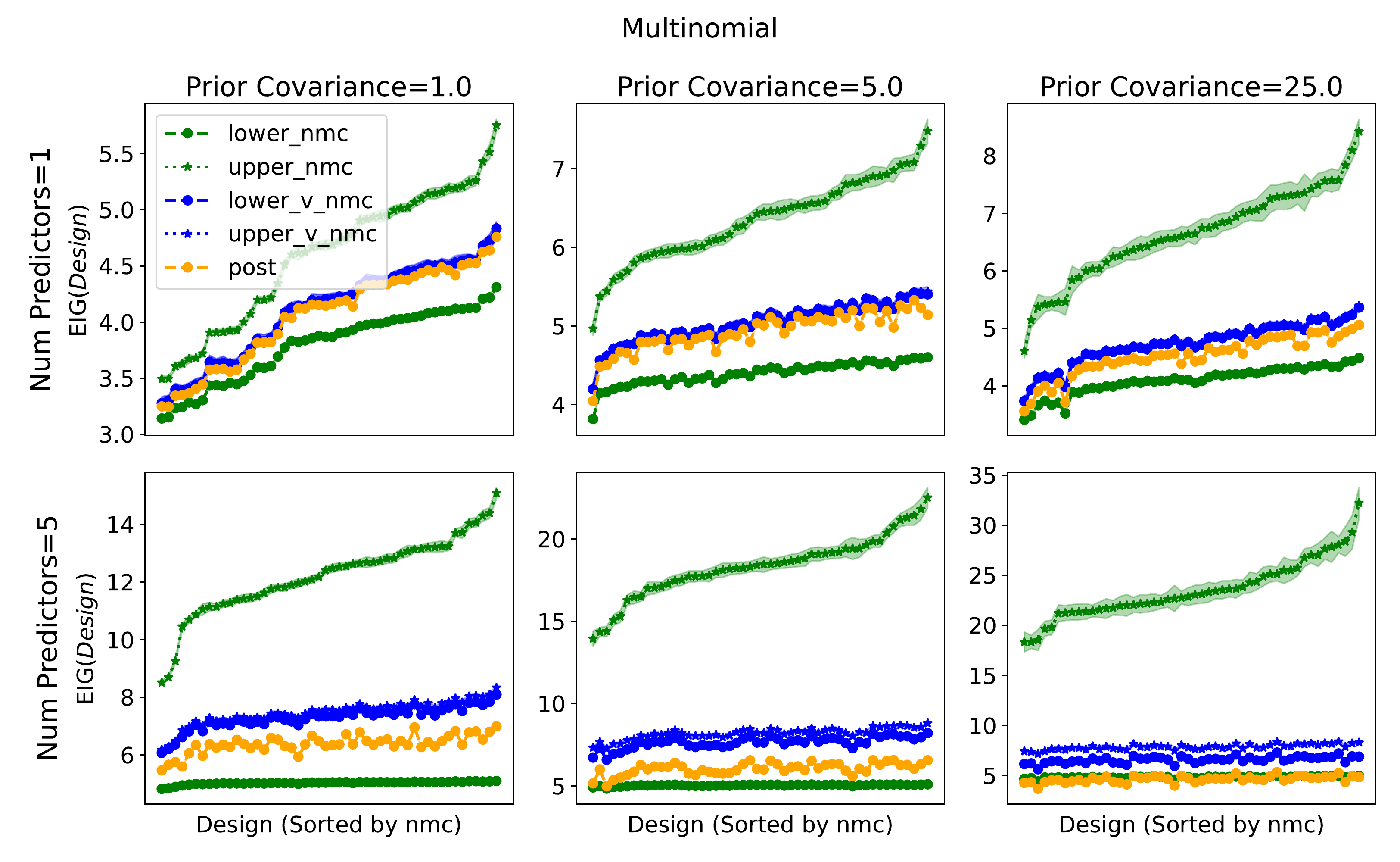}
    \caption{Same as figure \ref{fig:linear_model}, but for the multinomial model. For this model we cannot compute ground truth, but we can still infer superior performance for our methods by looking at how VNMC produces much tighter bounds, both lower and upper.}
    \label{fig:multinomial_model}
\end{figure*}

We include result plots from our Model Experiments on the logistic, binomial, categorical and binomial model in Figures \ref{fig:logistic_model}, \ref{fig:binomial_model}, \ref{fig:categorical_model}, \ref{fig:multinomial_model}, respectively. For discussion see the main text.

\subsection{Additional Architecture Experiment Results}

\begin{figure*}
    \centering
    \includegraphics[width=1.9\columnwidth]{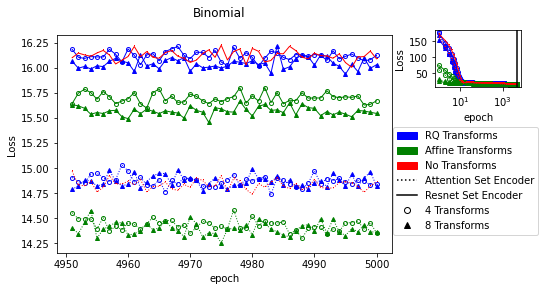}
    \caption{Same as figure \ref{fig:arch_loss_linear_unknown}, but for the binomial model.}
    \label{fig:arch_loss_binomial}
\end{figure*}

\begin{figure*}
    \centering
    \includegraphics[width=1.9\columnwidth]{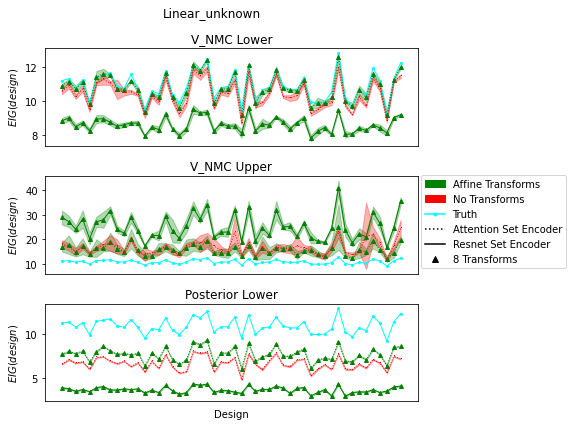}
    \caption{Here we show that the loss values in figure \ref{fig:arch_loss_linear_unknown} provide meaningful difference in the quality of EIG approximation. For a subset of the architectures trained we estimate the two VNMC based and the posterior bound using the same number of samples. We can see that the architectures that achieved lower loss values during training achieve greater estimation accuracy given the number of samples.}
    \label{fig:arch_eig_linear_unknown}
\end{figure*}

\begin{figure*}
    \centering
    \includegraphics[width=1.9\columnwidth]{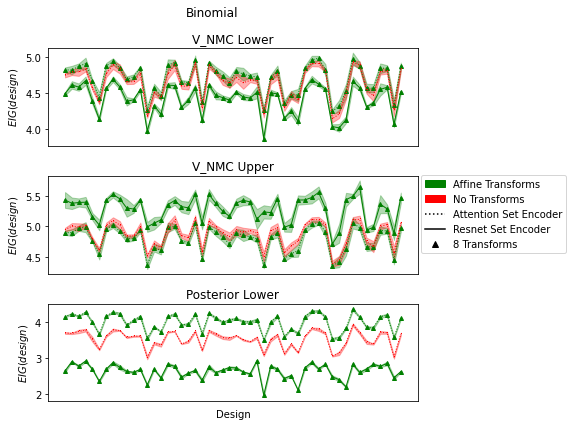}
    \caption{Same as in figure \ref{fig:arch_eig_linear_unknown}, but for the binomial model.}
    \label{fig:arch_eig_binomial}
\end{figure*}

Figure \ref{fig:arch_loss_binomial} shows the same plot as Figure \ref{fig:arch_loss_linear_unknown}, but for the binomial model. As discussed in the main text, we see that attention layers in the set encoder is the most important architectural decision, followed by the use of affine coupling layers. Figures \ref{fig:arch_eig_linear_unknown} and \ref{fig:arch_eig_binomial} show EIG estimates from a subset of the models trained in Figures \ref{fig:arch_loss_linear_unknown} and \ref{fig:arch_loss_binomial} for 50 randomly generated designs. We can see that models that achieved lower loss values are correspondingly more accurate for EIG estimation. See further discussion in main text.

\subsection{Full Architecture Details}
For the design amortization experiments and the model experiments in sections \ref{sec:amorization_experiments} and \ref{sec:model_experiments} we used a common neural network architecture across all models types, which we specify in detail for reproducibility. 
Note that, between all neural network layers described in the sequel is a ReLU activation function.

For a given experimental design matrix $D$ of size $N_{E} \times (N_{p}+1)$  we simulate from the model the expected outcomes $y$, a vector of size $N_{E}$. We concatenate these together to construct an input matrix, $C$,  of size $N_{E} \times (N_{p}+1 + 1)$ to our Set Invariant Model. Each row of this input matrix is then passed through an embedding network, which is a residual network with 2 residual blocks of dimension 64 (we define a residual block as 2 linear layers where the input to the first is added to the output of the second), creating an internal representation $R$ of size $N_{E} \times 120$. We now pass $R$ through 2 attention layers with 12 heads (head dimension is 10). Each attention layer is followed by a dropout layer, then a linear projection with 32 dimensions and another dropout layer; each dropout layer has a dropout probability 0.1.  This completes the set encoder, creating an internal representation for each experimental unit. These representations are then passed through the permutation invariant aggregator function, for which we use summation. 

Aggregation produces a single vector, regardless of the number of experimental units. This representation is then passed through the Emitter Network of the Set Invariant Model. The Emitter Network is simply a residual network with 2 residual blocks each with linear layers of dimension 128. This concludes the Set Invariant Model creating the design context $c_{y, d}$. 

We next provide the details of our conditional normalizing flow, following the sampling direction from the base distribution to the transforms. For the base distribution, we use a full rank multivariate normal distribution conditioned on the design context $c_{y, d}$. This distribution is parameterized by a residual network with 2 blocks and linear dimension of size 64. The last layer produces the mean vector of the normal distribution, and the entries of a lower triangular matrix that represents the Cholesky decomposition of the the covariance matrix. This lower triangular matrix is then left-multiplied with its transpose to creating the covariance of the base distribution. We can sample from this distribution straightforwardly. 

The samples are then passed through 4 conditional affine coupling layers (unless stated otherwise in the paper). Each affine coupling layer contains a residual neural network with two residual blocks with linear layers of dimension 128. The residual network takes as inputs the samples produced from the base distribution, concatenated with the context $c_{y, d}$. The network outputs the parameters for an affine transformation that is applied to the samples, which are then passed through a random permutation before the next affine coupling layer. The final affine coupling layer outputs the samples $\theta$ from our variational model $q_{\phi}(\theta |y, d)$. 

The forward procedure produces samples from our variational model, $q(\theta|y,d)$.  In order to evaluate $q(\theta|y,d)$ on samples simulated from the model, $p(y, \theta |d)$, we simply pass $y \text{and} d$ through the set invariant model to compute the design context, $c_{y, d}$, and then pass the simulated samples $\theta$ back through the inverse of the affine coupling layers to the base distribution, which can be evaluated. This now defines both directions of our variational model. 


This completes the architecture details used in the Model Experiments.  In the architecture experiments, we use the same architectural skeleton, but alter it as necessary.  Specifically, we test using a ResNet within the Set Encoder instead of the attention layers; this ResNet consists of 4 residual blocks with linear layers of dimension 64. We also test using rational quadratic coupling splines in place of the affine coupling layers; these spline transformations also use ResNets with 2 residual blocks of 128 dimensions for the linear layers which output the parameters of an RQ spline with 20 buckets and linear tails.

All models are trained with the AdamW optimizer with a learning rate of $5\times10^{-4}$ and $\beta_{0}=0.9$ and $\beta_{1}=0.999$ for 5000 steps, where each step consists of a batch of 50 randomly generated designs. During training we use the posterior estimator to define the loss, and set $N=50$ for the Monte Carlo estimator. We found no meaningful variation on the random seeds.